\documentclass[fleqn,10pt]{wlscirep}
\usepackage[utf8]{inputenc}
\usepackage[T1]{fontenc}
\usepackage{hyperref}
\hypersetup{
  hidelinks,
  pdfborder={0 0 0}
}
\usepackage{url}
\usepackage{microtype}
\usepackage{makecell}
\usepackage{graphicx}
\usepackage{subcaption}
\usepackage{csquotes}
\usepackage{longtable}
\usepackage{graphicx}
\usepackage{svg}

\usepackage{amsmath}
\usepackage{amssymb}
\usepackage{mathtools}
\usepackage{amsthm}
\usepackage{xspace}
\usepackage{colortbl}
\usepackage{multirow}
\usepackage{wrapfig}
\usepackage{float}
\usepackage{placeins}
\usepackage{listings}
\usepackage{color}
\usepackage{booktabs}
\usepackage{stackengine} 
\usepackage{fvextra}
\usepackage{tikz}
 
\usepackage[numbers,sort&compress]{natbib}

\usepackage{enumitem}
\setlist[itemize]{noitemsep}

\newcommand{\mname}{DrugAgent}

\title{\mname: Reliable Multi-Agent Integration of Conflicting Biomedical Evidence for Drug–Target Interaction Assessment}

\author[1,2]{Yoshitaka Inoue}
\author[1]{Tianci Song}
\author[3]{Xinling Wang}
\author[1]{Rui Kuang}
\author[5,+]{Tianfan Fu}
\author[2,4,+,*]{Augustin Luna}

\affil[1]{Department of Computer Science and Engineering, University of Minnesota, Minneapolis, MN, USA}
\affil[2]{Computational Biology Branch, National Library of Medicine, Bethesda, MD, USA}
\affil[3]{Khoury College of Computer Sciences, Northeastern University, Arlington, VA, USA}
\affil[4]{Developmental Therapeutics Branch, National Cancer Institute, Bethesda, MD, USA}
\affil[5]{State Key Laboratory for Novel Software Technology at Nanjing University, School of Computer Science, Nanjing University, Nanjing, Jiangsu, China}

\affil[+]{These authors jointly supervised this work.}
\affil[*]{augustin.luna@nih.gov}

\begin{abstract}
Workflows in drug-target interaction (DTI) assessment require integrating heterogeneous data from predictive models, curated resources, and observations from experimental literature. This evidence can be incomplete or conflicting. DrugAgent is a large language model (LLM)-based multi-agent system focused on DTI evidence integration that integrates outputs from machine learning, knowledge graph, and retrieval-augmented generation (RAG) agents. DrugAgent converts agent outputs into interpretable representations, then summarizes conflict across the evidence.

We evaluated DrugAgent on kinase screening data of 900 pairs spanning 178 kinases and 42 inhibitors, and an androgen receptor antagonist screening benchmark. On the kinase dataset, LLM-as-a-Judge evaluation indicated outputs were faithful to input evidence in 98.8\% of cases. Biological plausibility of returned summarization was high (scores 3-4 out of 5) across ground-truth classes: 79\% of Weak activity labels cases (81\% for Moderate/77\% Strong); Strong cases received higher scores than Weak/Moderate. Label stability showed 98\% agreement across runs. Results on the antagonist benchmark were consistent with the kinase dataset. 

Retrieved literature provided the greatest benefit when direct drug-target evidence was available, highlighting the importance of evidence availability for RAG-based integration. DrugAgent provides heterogeneous evidence-grounded DTI assessment, complementing standalone DTI prediction. We provide strategies to model agreement, conflict, and uncertainty in biomedical evidence integration. Code: \url{https://github.com/sciluna/DrugAgent}.
\end{abstract}
\begin{document}

\flushbottom
\maketitle

\thispagestyle{empty}

\begin{figure}[t]
\centering
\includegraphics[width=\textwidth,keepaspectratio]{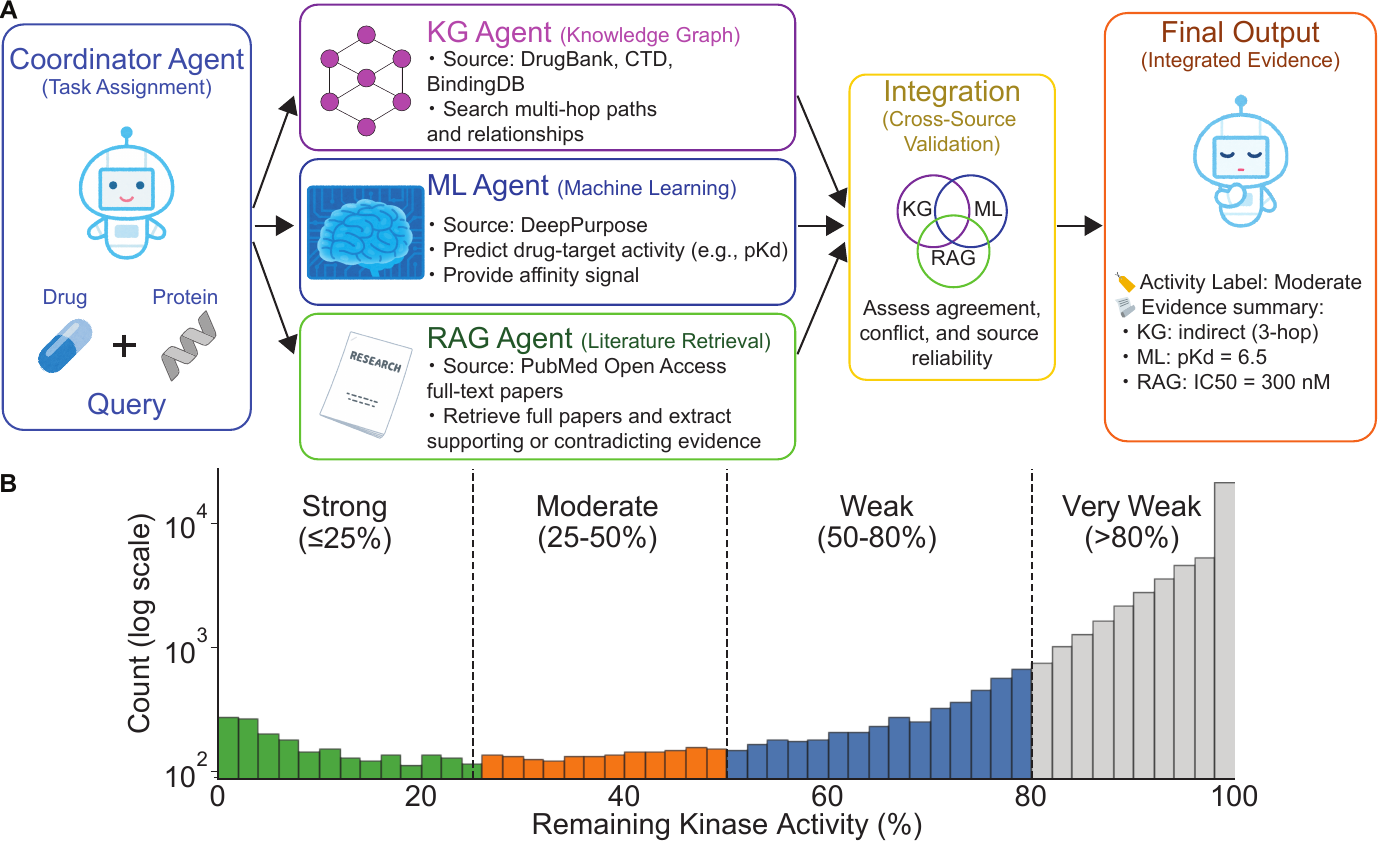}
\caption{(A) Overview of DrugAgent, a structured multi-agent framework for aggregating heterogeneous evidence in drug-target interaction (DTI) analysis. Three specialized agents independently generate complementary signals: a knowledge graph (KG) agent capturing mechanistic paths, a machine learning (ML) agent providing quantitative affinity predictions (e.g., pKd), and a literature retrieval (RAG) agent extracting experimental evidence. These signals are integrated through an integration step that enforces cross-source consistency. A reasoning module then performs structured evidence synthesis and decision-making, yielding a final prediction with an interpretable and reproducible rationale. (B) Histogram of remaining kinase activity in the kinase activity dataset, shown with log-scale counts on the y-axis. Samples are grouped into activity-based labels: Strong ($\leq 25\%$), Moderate ($25$-$50\%$), Weak ($50$-$80\%$), and very weak (>80\% remaining activity) are shown for context but were excluded from the balanced three-class kinase benchmark.}
\label{fig:overview}
\end{figure}

\section*{Introduction}

Drug-target interaction (DTI) assessment is a central task in early-stage drug discovery~\citep{abbasi2023drug, inoue2026review}. Reliable DTI assessment often requires integrating multiple forms of evidence, including quantitative predictions from machine learning (ML) models, curated drug-target knowledge from knowledge graphs (KGs), and experimental findings reported in the literature~\citep{huang2020deeppurpose,himmelstein2017systematic,lewis2020retrieval}. These evidence sources are complementary but imperfect. ML models can provide scalable affinity estimates, but they may be poorly calibrated outside their training distribution~\citep{ozturk2018deepdta,huang2020deeppurpose}. KGs organize curated biomedical relationships, but graph connectivity does not necessarily imply direct binding or functional modulation~\citep{himmelstein2017systematic}. Literature evidence can provide experimental context, but it is unevenly distributed across drugs and targets and may contain assay-specific or conflicting findings~\citep{stoeger2018large,turner2008selective}. As a result, DTI assessment is not only a prediction problem, but also an evidence integration problem.

In practice, researchers often need to reconcile information from distinct biological modalities (e.g., genomic vs proteomic), representation modalities (e.g., text, numeric, graph-based), and evidence provenances (e.g., observation, curated, or inference). Typically, researchers must manually check where evidence agrees, conflicts, or provides different levels of support. Large language models (LLMs) and multi-agent systems provide a possible interface for organizing disparate evidence~\citep{wei2022emergent,du2023improving}. Typical use of LLMs only includes access to knowledge available as part of their training data (usually text or image data) without access to other data sources that can improve performance across downstream tasks~\citep{schick2023toolformer}. Additionally, unconstrained LLM reasoning can obscure which evidence source supports the conclusion and may overstate weak or indirect evidence~\citep{agarwal2024faithfulness,pandit2025medhallu}. Therefore, DTI-focused systems require mechanisms that make additional sources of biomedical information accessible to LLM agents in a manner that can be assessed for evidence agreement and conflict across the varied sources.

Here, we introduce DrugAgent, a DTI-focused evidence integration system that integrates ML, KG, and retrieval-augmented generation (RAG)-based literature evidence for DTI assessment. DrugAgent uses specialized modules to generate source-specific evidence and converts its outputs into a shared structured representation. A reasoning module then summarizes agreement and conflict before applying rule-guided final decision-making. This approach aims to improve reliability, and prior work has also examined rule-guided reasoning in LLMs~\citep{zhou2025rulearena}. This design separates evidence summarization from final decision-making and enables inspection of how each evidence source contributes to the conclusion. 

We evaluate DrugAgent with two DTI assessment scenarios: the Anastassiadis kinase inhibition benchmark and the Tox21 androgen receptor (AR) antagonist screening benchmark~\citep{anastassiadis2011comprehensive,thomas2018us}. The objective of this study is to evaluate whether structured integration can support interpretable, evidence-grounded assessment under heterogeneous and potentially conflicting evidence; our goal is not to introduce a new standalone DTI predictor. Additionally, we assess biological plausibility, stability, and conflict behavior for reasoning returned from the agent system.

DrugAgent differs from prior DTI prediction tools that typically use a single mode of data (i.e., numeric data for predictive machine learning). By contrast, our multi-agent approach of treating DTI assessment as a structured evidence integration problem that jointly integrates ML predictions, KG-derived evidence, and RAG-based literature evidence into a shared representation to be integrated through LLM-based reasoning and summarization to final decision-making.

%this is similar to other areas of artificial intelligence that seek to integrate heterogeneous modalities remains a central challenge in artificial intelligence, despite substantial advances in large-scale multimodal models. ~\citep{genept, ye2024mplugowlmodularizationempowerslarge}

\section*{Related Works}

Human consideration of scientific results for DTI assessment often requires reconciling predictive algorithms, databases (e.g., knowledge graphs), and literature evidence that may differ in coverage, specificity, and reliability. We organize related work around these evidence sources and their integration for DTI analysis.

\subsection*{Machine learning for drug-target interaction prediction}

Models such as DeepDTA~\citep{ozturk2018deepdta} that use convolutional neural networks (CNN) and DeepPurpose~\citep{huang2020deeppurpose}, which includes several architectures such as CNN and message passing neural networks (MPNN) to predict drug-target activity from molecular structures and protein sequences. These models are valuable for scalable screening and provide quantitative affinity-related signals. However, their predictions are constrained by the coverage and biases of the training data, and they are typically optimized as standalone predictors rather than as systems for combining data from distinct biological modalities, representation modalities, and evidence provenances (e.g., observation, curated, or inference). As a result, they do not directly expose how their predictions should be reconciled with curated database evidence or experimental findings from the literature. In this study, we utilize a DeepPurpose-based ML module as one source of DTI evidence that has a numeric representation and is inferred.

\subsection*{Biomedical knowledge graphs for drug-target evidence}

Biomedical knowledge graphs integrate curated relationships among drugs, genes, diseases, and pathways with additional annotations such as experimental measurement values. Such data has been used for drug repurposing, link prediction, and mechanistic hypothesis generation~\citep{himmelstein2017systematic,drkg2020}. Knowledge graph embedding (KGE) methods such as TransE~\citep{bordes2013translating}, ComplEx~\citep{trouillon2016complex}, and RotatE~\citep{sun2019rotate} learn latent representations for predicting missing links from graph structure. These approaches can provide useful graph-derived relatedness scores. Still, their latent scores are not necessarily calibrated to assay-derived activity labels and may not expose the specific evidence paths supporting associations between drug and targets. Path-based methods, in contrast, preserve explicit graph connectivity and can provide interpretable evidence, but their implementation often depends on design choices such as maximum path length, hub-node handling, and edge-type selection~\citep{himmelstein2017systematic}. DrugAgent uses path-based KG evidence for traceability and reports TransE, ComplEx, and RotatE as graph-only reference baselines (Supplementary Table S3).

\subsection*{Literature evidence and retrieval-augmented biomedical analysis}

Retrieval-augmented generation (RAG) from text enables language models to condition their outputs on retrieved documents~\citep{lewis2020retrieval}. However, observations in scientific literature are unevenly distributed: well-studied drugs and targets may have abundant publications, whereas understudied pairs may have little or no direct evidence~\citep{stoeger2018large}. For DTI assessment, retrieved passages may provide direct drug-target pair evidence, one-sided drug or target context, or indirect mechanistic information, and these evidence types differ in reliability and specificity. DrugAgent therefore separates pair-level, drug-level, and target-level retrieval, allowing the system to distinguish direct interaction evidence from contextual or indirect support when aggregating evidence from text.

\subsection*{Multi-source evidence integration for DTI assessment}

A persistent challenge in DTI is how to combine heterogeneous modalities and provenance types when they agree, conflict, or provide different levels of support. ML predictions, KG-derived relationships, and RAG literature evidence can each contribute useful information, but none is uniformly reliable across all drug-target pairs. Recent work has examined methods for multi-agent systems to arrive at common final answers, while improving the accuracy of multi-agent debate in domains outside of biology~\citep{du2023improving,pitre2025consensagent,yao2025peacemaker}. DrugAgent addresses these problems by standardizing outputs from ML, KG, and RAG modules into a shared evidence representation and explicitly modeling agreement, conflict, and uncertainty before final decision-making. This design emphasizes interpretable evidence integration for DTI assessment rather than general-purpose multi-agent orchestration. Separately, but related to the integration of data from multiple sources, is the tool use by LLMs. This is an active area of investigation, both in terms of LLM capability to use tools and as a means to improve LLM performance~\citep{gpt5education, openai2025introducinggpt5, patil2025berkeley}. 

\section*{Datasets}

\subsection*{Kinase Inhibition Benchmark}

For quantitative evaluation, we used a protein inhibition kinase-compound activity dataset from a large-scale profiling study~\citep{anastassiadis2011comprehensive}. The dataset reports kinase activity as the percentage of remaining enzymatic function after compound exposure and includes 300 human kinases and 178 small-molecule inhibitors.

The original assay reports the percent remaining kinase activity after compound exposure. To formulate an activity label prediction, we discretized this continuous readout into three classes based on inhibition activity labels: Strong (remaining activity $\leq 25\%$), Moderate ($25$-$50\%$), and Weak ($50$-$80\%$), while pairs with $>80\%$ remaining activity were treated as very weak and excluded. These cutoffs are consistent with the activity ranges used to summarize kinase inhibition patterns in the original profiling study~\cite{anastassiadis2011comprehensive}. From 1,895 activity labels, we constructed a balanced benchmark of 900 pairs (300 per class), covering 178 kinases and 42 inhibitors.

\subsection*{Tox21 Androgen Receptor (AR) Dataset}
The Tox21 androgen receptor (AR) dataset was constructed from the Tox21 dataset annotation and reporter assay activity tables~\cite{thomas2018us}. We retained Tox21 AR reporter gene assays annotated with an antagonist mode of action. For each compound, reporter assay activity calls were represented as binary labels, with 1 indicating active and 0 indicating inactive. When multiple AR antagonist assay identifiers were available for a compound, their binary labels were collapsed by taking the maximum value across assays, such that a compound was labeled active for a gene if it was active in at least one antagonist assay.

Compound identifiers were linked to compound names using PubChem CIDs, and the resulting drug-target activity table was converted into drug-target pairs. To ensure that each instance could be evaluated using the evidence sources used by DrugAgent, we retained only compounds whose names matched entries in the knowledge graph. We then merged the activity table with the literature retrieval table and used the number of downloaded literature records for each drug-target pair as a proxy for evidence availability.

For the AR benchmark, we selected drug-AR pairs by ranking active and inactive pairs separately according to the number of downloaded literature records. The final balanced dataset consisted of 250 active and 250 inactive drug-AR pairs, yielding 500 total pairs.

\begin{table}[h]
\centering
\caption{
Summary of evaluation datasets. The kinase dataset is a controlled, class-balanced multi-class benchmark, while the Tox21 AR dataset is a binary screening benchmark.
}
\begin{tabular}{c|cc}
\toprule
 & Kinase Dataset & Tox21 (AR) Dataset \\
\midrule
Total pairs & 900 & 500 \\
Class distribution & 300 / 300 / 300 & 250 / 250 \\
Class labels & Weak / Moderate / Strong & Active / Inactive \\
Unique drugs & 42 & 500 \\
Unique targets & 178 & 1 (Androgen Receptor) \\
Assay type & Enzymatic inhibition & Receptor antagonism\\
Prediction task & Multi-class & Binary \\
\bottomrule
\end{tabular}
\end{table}

\section*{Results}
We evaluated DrugAgent along two complementary axes. First, we assessed whether the system produced evidence-grounded and internally consistent outputs when integrating ML, KG, and RAG evidence. This included LLMaJ-based diagnostics of faithfulness and biological plausibility, stability analysis across repeated runs, and analysis of integration behavior under agreement and conflict among evidence sources. Second, we performed sanity-check classification analyses on the kinase and Tox21 AR benchmarks to quantify whether the integrated outputs retained signal related to known activity labels. These evaluations were designed to characterize evidence integration behavior rather than to establish DrugAgent as an optimized standalone DTI predictor. Of note, DrugAgent produced short summaries with word counts ranging from 71 to 128 words across the agents for the tested benchmarks. 

\subsection*{Integration Fidelity}

\paragraph{(1) Faithfulness.}
We evaluated faithfulness using an LLMaJ rubric that scored explanations on a 1-5 scale according to whether they (i) were grounded to the outputs from each agent, (ii) contained contradictions between input evidence and output, (iii) introduced unsupported claims, or (iv) omitted critical information. The LLMaJ prompt also requested Boolean indicators for these four criteria. We defined outputs with faithfulness scores of 4 or 5 as faithful. Under this criterion, faithfulness was high across all configurations: DrugAgent was judged faithful in 889 of 900 cases (98.8\%), compared with 896/900 for KG+RAG, 857/900 for ML+RAG, and 900/900 for ML+KG.

\paragraph{(2) Biological Plausibility under a Prompted Rubric.}

\begin{figure}[!thb] \centering \includegraphics[width=\linewidth]{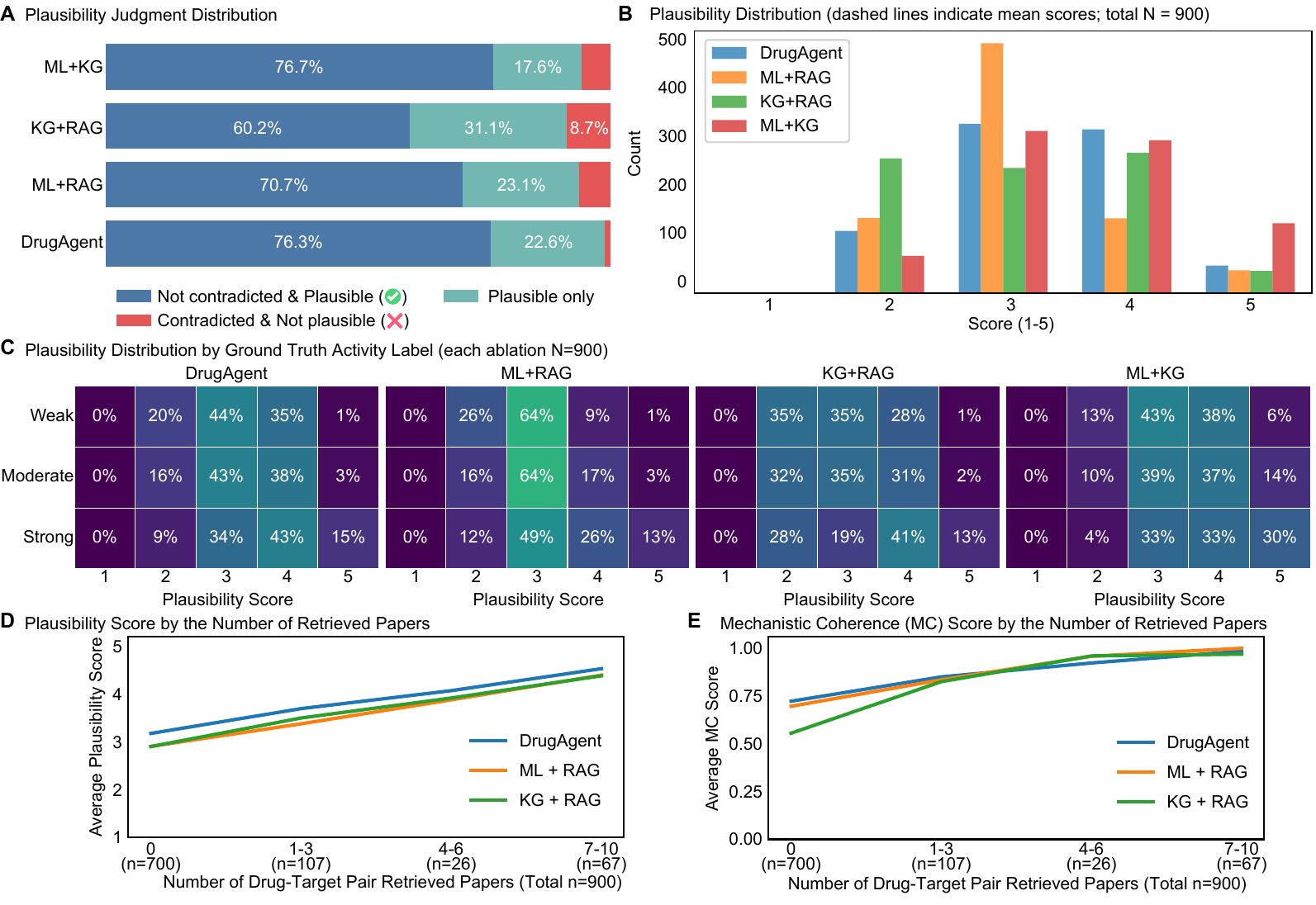} \caption{
Plausibility-based evaluation across model variants for the kinase dataset. (A) Distribution of LLMaJ plausibility outcomes for DrugAgent and ablated variants, showing the proportions of Not contradicted \& Plausible, Plausible only, and Contradicted \& Not plausible. (B) Distribution of plausibility scores (1-5) across variants; dashed vertical lines indicate the mean score for each variant. (C) Plausibility score distributions stratified by ground-truth activity labels (Weak, Moderate, Strong). Each heatmap shows the proportion of samples assigned to each plausibility score within a ground-truth class. (D) Average plausibility score across bins of the number of retrieved papers for each drug-target pair among RAG-containing settings. (E) Average mechanistic coherence (MC) score across bins of the number of retrieved papers for each drug-target pair among RAG-containing settings. 
}\label{fig:plausibility} \end{figure}

DrugAgent produced highly plausible and largely non-contradictory outputs under the prompted LLMaJ rubric (Fig.~\ref{fig:plausibility}A). Overall, 76.3\% of DrugAgent outputs were judged both not contradicted and plausible, and an additional 22.6\% were judged plausible only. Thus, 98.9\% of DrugAgent outputs were considered at least plausible, while only a small fraction were rated as contradicted and not plausible. Among the ablations, ML+KG showed a similar proportion of not-contradicted and plausible associations in the outputs (76.7\%), whereas ML+RAG and KG+RAG showed lower values of 70.7\% and 60.2\%, respectively. Notably, removing ML increased the ``contradicted and not plausible'' fraction to 8.7\%, indicating that the ML component contributes to filtering biologically inconsistent conclusions.

The score distribution further shows that plausibility scores should not be interpreted as direct evidence of predictive correctness (Fig.~\ref{fig:plausibility}B). DrugAgent scores were concentrated at 3 and 4, with relatively few scores of 5. In contrast, ML+KG shifted the distribution toward higher scores, including a larger number of score-5 judgments. However, this increase occurred despite removing retrieved literature evidence from the integration process. Therefore, higher plausibility scores in the ML+KG setting may reflect reduced evidential constraint or more fluent surface-level coherence, rather than stronger evidence-grounded reasoning.

Stratification by ground-truth activity labels showed that plausibility scores were only weakly aligned with the actual ordinal activity labels (Fig.~\ref{fig:plausibility}C). In DrugAgent, most Weak, Moderate, and Strong interactions received scores in the 3-4 range. Specifically, 79\% of Weak cases, 81\% of Moderate cases, and 77\% of Strong cases were assigned scores of 3 or 4. Strong interactions were more likely to receive the maximum score of 5 than Weak or Moderate interactions (15\% for Strong versus 1\% for Weak and 3\% for Moderate), but the overall distributions remained highly overlapping. 

We further examined whether the amount of retrieved evidence was associated with judge-assigned plausibility. Across all variants, the average plausibility score increased with the number of retrieved papers for a drug-target pair (Fig.~\ref{fig:plausibility}D). Pairs with no retrieved papers received the lowest average scores, whereas pairs supported by 7-10 retrieved papers received the highest scores. DrugAgent consistently achieved the highest average plausibility score across evidence-count bins, suggesting that its retrieved context yielded more biologically plausible associations than the ablated variants.

A similar trend was observed for mechanistic coherence (Fig.~\ref{fig:plausibility}E). The average mechanistic coherence (MC) score increased as the number of retrieved papers increased, with all variants approaching high coherence scores when multiple supporting papers were retrieved. Differences among ablation conditions were most apparent when little or no literature evidence was retrieved, whereas MC converged across variants for pairs with larger numbers of retrieved papers. These results indicate that richer retrieved evidence was associated with higher overall biological plausibility and MC of the inferred drug-target associations. 

Together, DrugAgent has less contradicted and not plausible portion (Fig~\ref{fig:plausibility}A), while judge-assigned plausibility remains strongly influenced by the availability of evidence retrieved by the agents (Fig~\ref{fig:plausibility}D). Plausibility scores were not directly predictive of ground-truth activity labels, although there is a trend toward this pattern (Supplementary Fig. S1); this could be because plausibility is influenced by other available evidential support. The ablation results indicate that individual evidence sources provide complementary constraints: ML mostly helps suppress biological plausibility and not contradiction, followed by KG (Fig~\ref{fig:plausibility}A), and RAG anchors explanations to retrieved literature (Fig~\ref{fig:plausibility}D). 

MC showed a similar evidence-dependent pattern, increasing with retrieved-paper count and converging across variants when multiple supporting papers were available. At the same time, the upward shift in ML+KG plausibility scores highlights a limitation of LLMaJ plausibility evaluation: less constrained explanations can appear more plausible even when they are less grounded in evidence.

\paragraph{(3) Stability.}
To evaluate reproducibility, we conducted a stability analysis on a randomly sampled subset of $N=100$ drug-target pairs, running each case three times ($k=3$). Label stability, defined as the fraction of cases with identical outputs across runs, shows a mean agreement of $0.983$, a median of $1.000$, and an agreement rate of $0.95$ where all outputs have the same label, indicating near-deterministic behavior with only minor variability across runs. Reasoning stability, measured by cosine similarity of embeddings between explanations, yields a mean similarity of $0.881$, a median of $0.884$, and a minimum of $0.760$, indicating that explanations remain highly consistent despite minor variations in phrasing. Together, these results demonstrate that the integration process is both label-stable and reasoning-consistent, supporting the reproducibility of the framework.

\paragraph{(4) Evidence Combination Consistency and Conflict Dynamics.}

% To analyze how heterogeneous evidence is integrated, we examine integration behavior conditioned on combinations of module outputs (ML, KG, RAG), representing each case as a triplet (e.g., M|W|S). KG-only agent has more ``Weak'' labels, while ML-only agent is dominated by ``Moderate'' labels. RAG-only and DrugAgent have more equal distributions (Figure).

\begin{figure}[!thb]
\centering
\includegraphics[width=\linewidth]{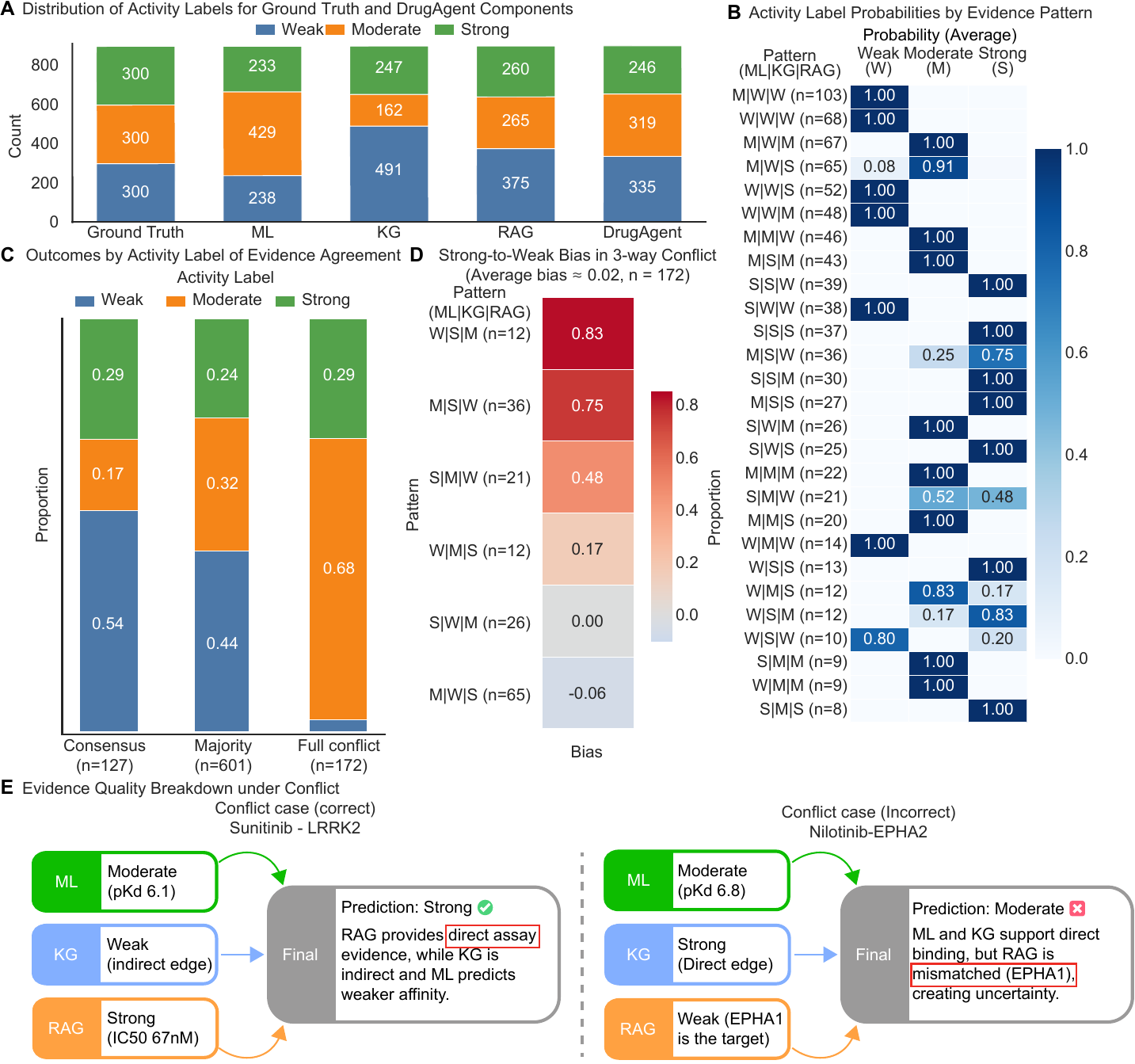}
\caption{
Label-level decision behavior under multi-source evidence integration for the kinase dataset.
(A) Bar plot showing activity labels for ground truth, and each of the DrugAgent components, as well as the final DrugAgent output.
(B) Heatmap of DrugAgent output probabilities for each ML$\vert$KG$\vert$RAG evidence-label pattern.
(C) Distribution of DrugAgent output labels by evidence-source agreement level, including consensus, majority-agreement, and full-conflict cases.
(D) Directional bias under full conflict, defined as $P(\mathrm{Strong}) - P(\mathrm{Weak})$, for each conflicting evidence pattern.
(E) Representative conflict cases showing successful and failed integration; red highlights indicate key text related to label decision.
}
\label{fig:combination_distribution}
\end{figure}

Figure~\ref{fig:combination_distribution} shows different aspects of label-level distribution after multi-source evidence integration. The bar diagram in Fig.~\ref{fig:combination_distribution}A shows the activity labels for ground truth, and each component of DrugAgent, as well as the final DrugAgent output. Although the ground-truth set was balanced across Strong, Moderate, and Weak labels, the individual agent sources showed distinct redistribution patterns. ML predictions have a larger number of Moderate assignments, whereas the KG agent results in a larger fraction of Weak labels. RAG evidence was more evenly distributed, with a slight shift toward Weak and Moderate labels. After integration, DrugAgent results in 246 pairs to Strong, 319 to Moderate, and 335 to Weak, indicating that the final decision process does not simply mimic any single evidence source but instead has a unique distribution of the three activity labels.

We next quantified how specific combinations of ML, KG, and RAG evidence labels were mapped to final DrugAgent outputs. 
In cases with completely concordant agent evidence, DrugAgent assigned the corresponding final label: patterns such as W$\vert$W$\vert$W, M$\vert$M$\vert$M, and S$\vert$S$\vert$S were mapped entirely to Weak, Moderate, and Strong outputs, respectively (Fig.~\ref{fig:combination_distribution}B). 
When the evidence sources disagreed, DrugAgent outputs varied according to the specific configuration of ML, KG, and RAG labels rather than following a uniform averaging pattern. 
For example, patterns with two equivalent DrugAgent labels (e.g., W$\vert$W$\vert$S) were assigned to the most common label except the W$\vert$S$\vert$W label, although even in this case, Weak was the most commonly assigned label. These results indicate that DrugAgent's final activity labels are strongly conditioned on the joint evidence pattern across ML, KG, and RAG.

As shown in Fig.~\ref{fig:combination_distribution}C, we next grouped cases by the degree of agreement among the three evidence sources. 
In cases where all evidence sources were in consensus, DrugAgent assigned 54\% of cases to Weak, 17\% to Moderate, and 29\% to Strong. 
In majority-agreement cases, the final labels were more broadly distributed, with 44\% Weak, 32\% Moderate, and 24\% Strong assignments. 
Under full conflict, DrugAgent resulted in more Moderate labels, assigning 68\% of cases to Moderate and 29\% to Strong, with only a small fraction assigned to Weak.
These results indicate that DrugAgent's integration behavior depends on the level of agreement among ML, KG, and RAG evidence sources, with fully conflicting evidence tending to produce more moderate final decisions.

We further examined whether DrugAgent showed systematic directional bias when all three evidence sources disagreed. 
As shown in Fig.~\ref{fig:combination_distribution}D, three-way conflicting patterns produced non-uniform final-label biases when comparing Strong to Weak labels.
Several patterns showed positive bias, indicating that DrugAgent outputs were skewed toward Strong rather than a Weak final label.
The strongest bias was observed for W$\vert$S$\vert$M and M$\vert$S$\vert$W patterns, where there was a Strong label produced by the KG agent. By contrast, S$\vert$W$\vert$M showed no positive bias and M$\vert$W$\vert$S showed a slight negative bias toward the Strong label.
The average bias across three-way conflict cases was close to zero (Fig.~\ref{fig:combination_distribution}D), suggesting that DrugAgent did not globally favor stronger or weaker labels under conflict, although individual conflict patterns showed clear directional differences.

Finally, we inspected representative drug-target pairs to illustrate how DrugAgent behaved under evidence disagreement. 
DrugAgent recovered the ground-truth activity label despite discordant ML, KG, and RAG evidence, in 43\% of cases (74 out of 172 cases), where there was a 3-way conflict of individual agent labels (Fig.~\ref{fig:combination_distribution}E). 

Examples in Fig.~\ref{fig:combination_distribution}E indicate evidence labels or final outputs that disagree with the ground-truth label. Text in the boxes labeled ``Final'' summarizes the reasoning used by DrugAgent. The phrases highlighted in red show elements that aided or harmed the ability of DrugAgent to make the correct label assignment; for the left-hand example in the figure, the observation of direct quantitative measurements gave DrugAgent reasoning confidence to produce a ``Strong'' label, while in the right-hand example, the lack of drug-target co-mention led to uncertainty. These examples demonstrate that DrugAgent can integrate heterogeneous evidence to preserve or recover the correct label in some conflict settings, but its integration is not uniformly reliable when evidence sources disagree.

\paragraph{Use case (2) using the Tox21 AR Screening Dataset.}

\begin{figure}[!thb]
\centering
\includegraphics[width=\linewidth]{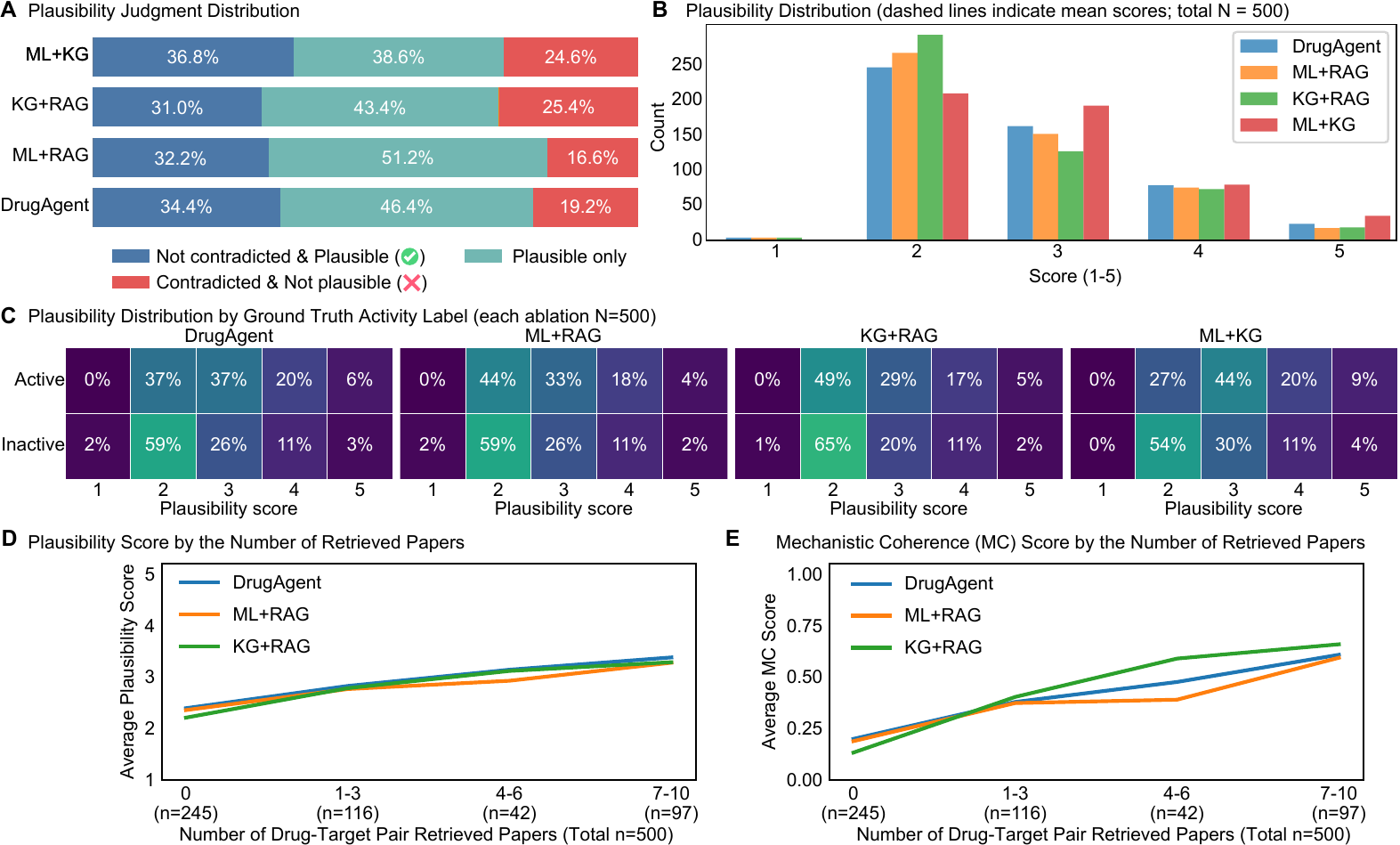}
\caption{
Plausibility-based evaluation across model variants for the Tox21 dataset. (A) Distribution of plausibility judgment states across the Tox21 AR dataset, showing the proportions of Not contradicted \& Plausible, Plausible only, and Contradicted \& Not plausible across DrugAgent and ablated variants. 
(B) Distribution of plausibility scores across ablation settings; dashed vertical lines indicate the mean score for each variant. 
(C) Plausibility score distributions stratified by assay label. The heatmaps compare Active and Inactive compounds and assess whether evidence integration reflects assay-derived activity labels.
(D) Mean plausibility score across bins of the number of retrieved papers for each drug-target pair among RAG-containing settings.
(E) Mean mechanistic coherence score across bins of the number of retrieved papers for each drug-target pair among RAG-containing settings.
}
\label{fig:tox21}
\end{figure}

The Tox21 AR dataset was used as a use case with binary class labels (Fig.~\ref{fig:tox21}). DrugAgent produced a fraction of outputs judged ``Not Contradicted \& Plausible'' (34.4\%), while removing KG increased the fraction of ``Plausible only'' outputs to 51.2\%, suggesting that KG removal weakened the evidence needed for outputs to satisfy both non-contradiction and plausibility criteria (Fig.~\ref{fig:tox21}A). In addition, compared with DrugAgent, ML+KG increased the fraction of ``Not Contradicted \& Plausible'' outputs by 2.4 percentage points, but also increased the fraction of ``Contradicted \& Not plausible'' outputs by 5.4 percentage points. Overall, ML+KG shifted the plausibility score distribution upward (mean $\simeq 2.88$), with more judgments assigned to scores of 3 or higher despite the increased contradiction rate (Fig.~\ref{fig:tox21}B).

DrugAgent plausibility scores showed differences between Active and Inactive compounds, with Active samples having higher scores than Inactive samples. In DrugAgent, 63\% of Active cases received scores of 3-5, compared with 40\% of Inactive cases, whereas Inactive cases were more concentrated at score 2 (59\% vs. 37\% for Active). This difference appeared weaker when ML or KG was removed, suggesting that molecular predictive signals and structured knowledge can contribute to activity-related plausibility patterns in this dataset (Fig.~\ref{fig:tox21}C). As such, one observation regarding the plausibility scores reported for DrugAgent (or any of the tested ablations) is that these scores did not necessarily show an interdependence with the assay-derived labels. This was evidenced by the small percentages of low and high plausibility scores (scores either 1 or 5) throughout the ablations. This pattern was observed in both the kinase and Tox21 datasets (Fig.~\ref{fig:plausibility}C and Fig.~\ref{fig:tox21}C). Readers are reminded that plausibility scores are generated from evaluation via LLMaJ. We also observe that the ML+KG ablation setting can result in high plausibility scoring for the Strong classification label for the kinase dataset (Fig.~\ref{fig:plausibility}C). This suggests that despite excluding retrieved literature, evidence from only ML+KG can be sufficiently clear without being externally grounded (i.e., through publication text).   

Average plausibility scores generally increased with the number of retrieved papers, although the trend was modest and not strictly monotonic across all variants (Fig.~\ref{fig:tox21}D). Mechanistic coherence (MC) showed a clearer evidence-dependent pattern, with MC scores increasing from low values in the no-retrieval bin to the highest values for pairs supported by 7-10 retrieved papers (Fig.~\ref{fig:tox21}E). These results suggest that literature availability is associated with stronger judge-assigned biological plausibility and, more prominently, MC in the Tox21 AR setting; this is similar to the kinase dataset use case.

\subsection*{Sanity-Check DTI Classification}

\begin{figure}[!thb]
\centering
\includegraphics[width=\linewidth]{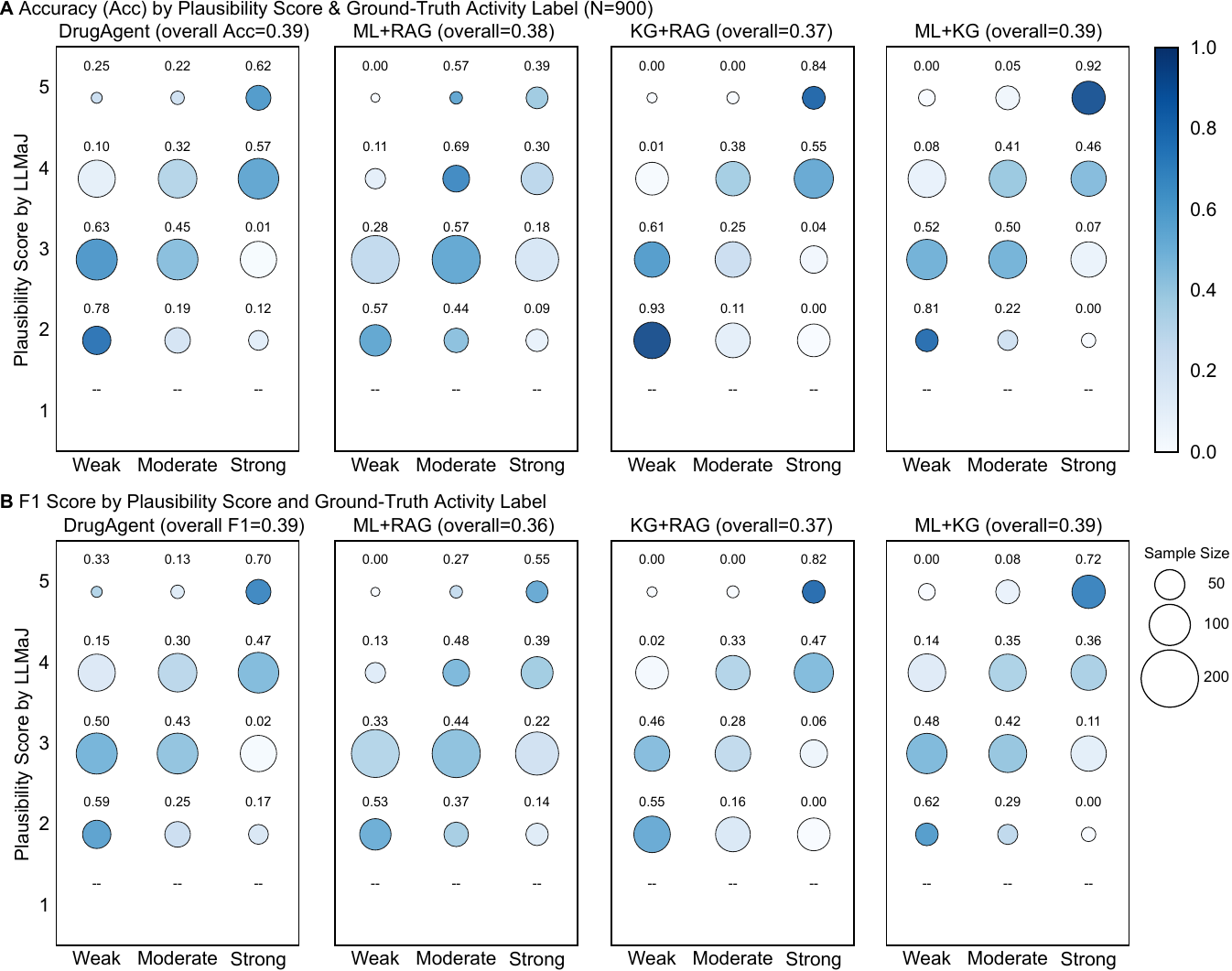}
\caption{
Stratified predictive performance across LLMaJ plausibility scores and ground-truth activity labels on the kinase benchmark.
(A) Accuracy and (B) F1 score for DrugAgent and ablated model variants. Rows represent plausibility scores assigned by the LLMaJ, ranging from 5, highest plausibility, to 1, lowest plausibility. Columns represent ground-truth activity labels. Circle color encodes the performance metric, and circle area encodes the number of samples per cell. Cells without observations are shown as empty.
}
\label{fig:kinase_pred}
\end{figure}

\begin{figure}[!thb]
\centering
\includegraphics{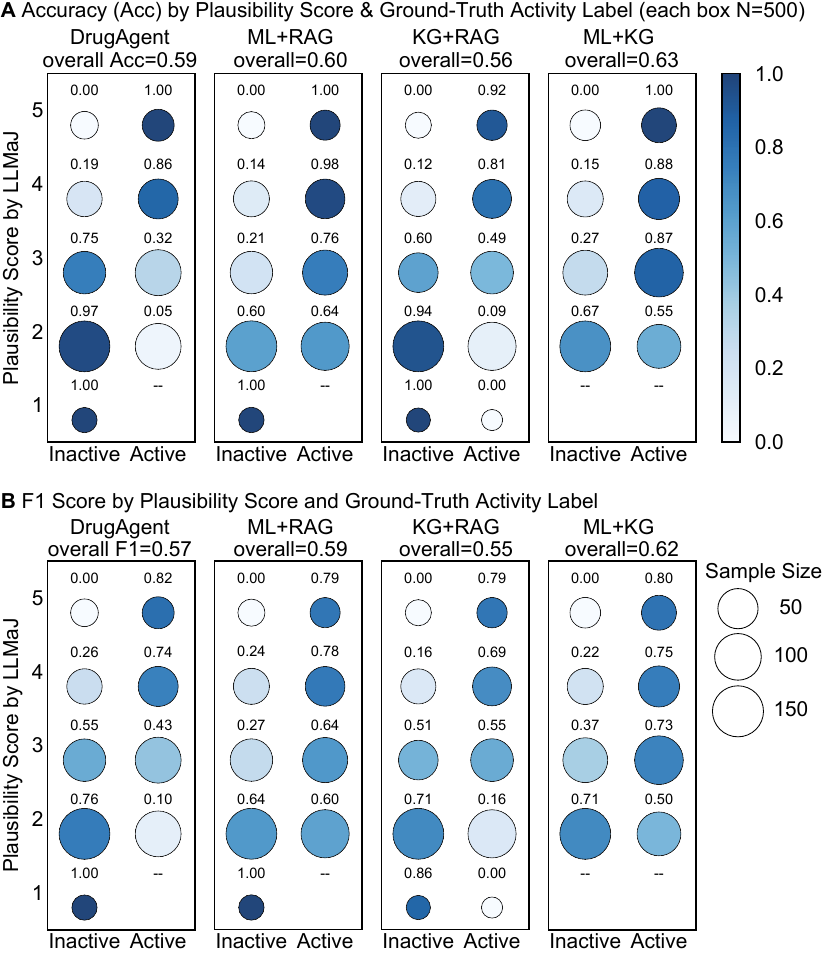}
\caption{
Model performance across LLMaJ plausibility strata and Tox21 activity labels.
(A) Accuracy and (B) F1 score were computed within each stratum defined by LLMaJ plausibility score and ground-truth activity label. Columns compare DrugAgent with ablated variants excluding KG, ML, or RAG components. Bubble color denotes the metric value, bubble area denotes the number of samples in each stratum, and numeric annotations indicate the corresponding metric values.
}
\label{fig:tox21_pred}
\end{figure}

To assess whether LLM-as-a-Judge (LLMaJ) plausibility scores were aligned with known DTI labels, we performed a sanity-check classification analysis on the kinase and Tox21 benchmarks(Fig.~\ref{fig:kinase_pred} and Fig.~\ref{fig:tox21_pred}). The kinase benchmark used ordinal activity labels (Weak, Moderate, and Strong), whereas the Tox21 benchmark used binary assay labels (Inactive and Active). For each benchmark, we stratified predictions by LLMaJ plausibility score and ground-truth label, and computed accuracy and F1 within each stratum for DrugAgent and the ablated variants.

This analysis was intended to test whether higher plausibility judgments corresponded to better agreement with known labels, rather than to treat plausibility as a direct measure of predictive correctness. Overall performance was summarized using accuracy and F1, while stratum-level values were used to identify where plausibility scores did or did not align with label-specific predictive performance. For reference, a uniform random classifier would have an expected accuracy of 0.33 for the three-class kinase benchmark and 0.50 for the binary Tox21 benchmark.

In the kinase dataset, overall performance was modest and similar across variants (Fig.~\ref{fig:kinase_pred}). DrugAgent achieved an overall accuracy of 0.39 and F1 of 0.39, comparable to ML+KG and slightly higher than ML+RAG and KG+RAG. DrugAgent showed improvement over ML+KG in conditions such as low/high plausibility (score-3 and 5) Weak entries as well as low/moderate (score-2 and 4) plausibility Strong entries. For example, high plausibility (score-4) Strong examples (accuracy = 0.57; F1 = 0.47; n = 128). DrugAgent also showed relatively high performance for score-3 Weak examples (accuracy = 0.63; F1 = 0.50; n = 131).

In the Tox21 dataset, DrugAgent achieved an overall accuracy of 0.59 and a support-weighted F1 of 0.57 (Fig.~\ref{fig:tox21_pred}). Similar to the kinase dataset,  DrugAgent performed strongly for low plausibility (score-2) Inactive examples (accuracy = 0.97; F1 = 0.76; n = 147); this is consistent with the expected relationship between low plausibility and inactive assay labels. Separately, DrugAgent performed well for score-3 Inactive examples (accuracy = 0.75; F1 = 0.55; n = 65). Additionally, each ablation condition for DrugAgent had high accuracy for high plausibility (scores 4 and 5) Active entries.

\begin{figure}[t]
\centering

\begin{minipage}[t]{0.46\textwidth}
\vspace{0pt}
\centering
\small
\resizebox{\linewidth}{!}{%
\begin{tabular}{c|cc|cc}
\toprule 
Model & \multicolumn{2}{c|}{Kinase dataset} & \multicolumn{2}{c}{Tox21 dataset} \\
 & Acc & F1 & Acc & F1 \\
\midrule
 ML only & 0.37 & 0.37 & 0.61 & 0.60 \\
 KG only & 0.38 & 0.35 & 0.58 & 0.54 \\
 RAG only & 0.34 & 0.33 & 0.52 & 0.48 \\
\midrule
 ML+KG  & \textbf{0.39} & \textbf{0.39} & 0.63 & 0.62 \\
 ML+RAG & 0.38 & 0.36 & 0.60 & 0.59 \\
 KG+RAG & 0.38 & 0.36 & 0.56 & 0.55 \\
\midrule
 DrugAgent & \textbf{0.39} & \textbf{0.39} & 0.59 & 0.57 \\
 \midrule
\makecell[c]{DrugAgent\\(Skip RAG without DTI papers)}
& \textbf{0.39}
& \textbf{0.39}
& \textbf{0.67}
& \textbf{0.67} \\
\bottomrule
\end{tabular}%
}
\captionof{table}{Ablation performance across standalone components, leave-one-component-out settings, and the full DrugAgent framework. Accuracy (Acc) and F1 are reported for the Kinase and Tox21 datasets. Bold values indicate the best performance within each dataset and metric.}
\label{tab:ablation_full}
\end{minipage}
\hfill
\begin{minipage}[t]{0.52\textwidth}
\vspace{0pt}
\centering
\includegraphics[width=\linewidth]{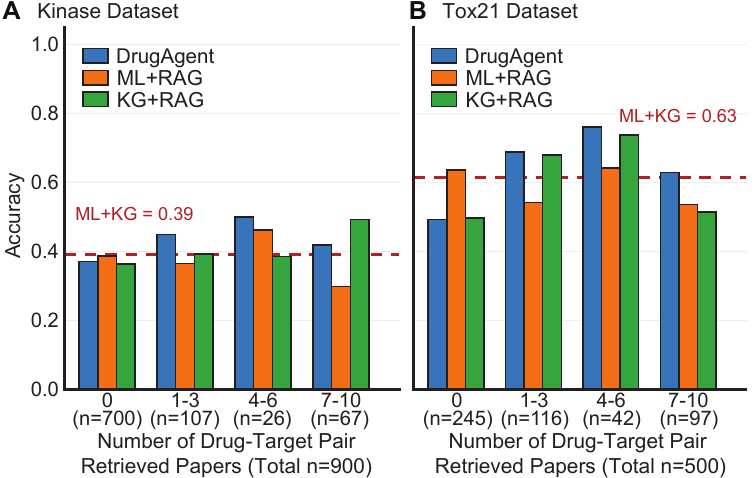}
\captionof{figure}{Accuracy stratified by the number of retrieved papers for (A) Kinase and (B) Tox21 AR datasets. Bars show the performance of DrugAgent and its RAG-based ablations, ML+RAG and KG+RAG, while the dashed red line indicates the ML+KG baseline.}
\label{fig:retrieval_acc}
\end{minipage}

\end{figure}

Although ML+KG achieved the highest accuracy and F1 on Tox21 and tied with DrugAgent on the kinase benchmark at 0.39 accuracy and 0.39 F1 score (Table~\ref{tab:ablation_full}), this advantage was not uniform across evidence-availability strata when accuracy was stratified by the number of retrieved papers (Fig.~\ref{fig:retrieval_acc}). DrugAgent and the RAG-based ablations showed improved performance as more literature evidence was available. In the kinase benchmark, DrugAgent exceeded the ML+KG baseline for all retrieved-paper count bins (from 1 to 10 publications) whereas performance was closer to or below the baseline when no papers were retrieved (Fig.~\ref{fig:retrieval_acc}A). A similar pattern was observed in Tox21, where DrugAgent exceeded the ML+KG baseline when at least one paper was retrieved and achieved its highest accuracy in the 4-6 retrieved-paper count bin (Fig.~\ref{fig:retrieval_acc}B). These results suggest that the apparent aggregate advantage of ML+KG is partly driven by evidence sparsity, while retrieved literature becomes beneficial when sufficient supporting evidence is available. 

These results suggest that selectively using RAG is beneficial. In particular, while all three types of papers (e.g., drug-only, gene-only, and drug-target pair papers) are always retrieved, the best overall performance can be achieved by a strategy that uses DrugAgent only when drug-target pair papers are available and otherwise uses ML+KG when no such papers are found. (Table~\ref{tab:ablation_full}). This strategy matches the best performance on the kinase dataset (accuracy = 0.39; F1 = 0.39) and substantially improves performance on the Tox21 dataset (accuracy = 0.67; F1 = 0.67).

We additionally evaluated baseline methods for the task of link prediction using knowledge graph embedding (KGE) models (i.e., TransE, ComplEx, and RotatE), as well as supervised predictive models (Supplementary Table S3 and Supplementary Table S4, respectively). Standard KGE methods showed limited alignment with assay-derived labels with performance ranging from 0.28 to 0.31 on the kinase dataset and reaching at most 0.57 accuracy and 0.57 F1 on Tox21 AR (Supplementary Table S3). This analysis was conducted under a leakage-controlled graph setting (Supplementary Methods). Next, we used several pretrained DeepPurpose models (each with a different architecture and dataset, including one with a CNN-based architecture similar to DeepDTA), which showed modest performance, reaching at most 0.37 accuracy/F1 on the kinase dataset and 0.61/0.60 accuracy/F1 on Tox21 AR (Supplementary Table S4). 

These results should therefore be interpreted with caution when DrugAgent using only KG as compared to existing KG link-prediction algorithms, because KG-only usage in DrugAgent is not meant to predict new links. Similarly, direct comparisons with pretrained ML results should be made cautiously, as such pretrained components cannot be controlled for data leakage.

\section*{Discussion}

This study addresses a central challenge in biomedical research, which is the integration of data with distinct biological modalities, representation modalities, and evidence provenances (e.g., observation, curated, or inference). There are additional challenges in the integration of this as scientific data can be incomplete or conflicting. A major aspect of this study to address this challenge is the usage of LLM-based agents (i.e., ML, KG, or RAG) for structuring and reconciling these different sources of evidence during integration. Rather than serving as a standalone drug-target interaction (DTI) predictor, DrugAgent is designed as a complementary tool for evidence integration. 

DrugAgent converts ML predictions, path-based KG evidence, and retrieved literature evidence into a common data structure suitable for LLM-based reasoning to assess the usage of such distinct data by LLMs for uncertainty and then summarize this data. Consistent with this objective, DrugAgent showed high faithfulness to the provided content, stable output behavior, and the ability to produce interpretable summaries of individual agent results, as well as reasoning logic for included final labels. 

\subsection*{Limitations and Future Work}

This work has several limitations. First, the evaluation was designed to characterize evidence integration behavior rather than strict predictive generalization. Although we evaluated DrugAgent on both a kinase inhibition benchmark and a Tox21 androgen receptor antagonist benchmark, we did not perform broader generalization checks such as zero-shot evaluation, temporal splits, cross-dataset benchmarking using additional datasets such as DAVIS or KIBA, or target-class-level evaluations across GPCRs, ion channels, transporters, and nuclear receptors beyond the androgen receptor. Therefore, further work is needed to assess performance for additional target classes and under unseen-drug, unseen-protein, temporal, dataset-level, target-class-level, and modality-level splits. Second, although the KG component uses interpretable path-based evidence useful for reasoning, its current scoring procedure is heuristic and relies on fixed constraints related to path length and hub-node penalties.

Additionally, DrugAgent does not currently learn calibrated integration weights for evidence sources; automated and systematic review of academic research results remains a challenge in the artificial intelligence field~\citep{dycke-gurevych-2026-automatic,ferrer2026trustllmgraderscalibrating}. Related to this, our RAG agent is constrained by the coverage and biases of the indexed literature. This may introduce publication bias, as well-studied proteins and drugs are more likely to be represented by numerous publications~\cite{stoeger2018large}. Moreover, although the retrieval step uses assay-related query terms, experimental context, assay type, and evidence strength are not explicitly encoded as structured metadata during retrieval; this is a similar issue experienced in many pathway interaction databases~\cite{giudice2024phuego}. 

Separately, DrugAgent integration remains rule-guided rather than statistically calibrated. Multi-agent execution is also more computationally expensive than single-model inference because ML scoring, KG path retrieval, RAG evidence retrieval, and LLM-based evidence synthesis introduce additional latency. The current implementation is therefore not optimized for high-throughput screening over very large drug-target matrices, and is better suited for evidence aggregation and interpretation of selected candidate pairs. Future work should explore improved source calibration and methods for improving cost and latency efficiency. 

There is evidence that verbose LLM explanations may inflate human confidence in the explanations provided, regardless of their factual correctness~\citep{zhao2026wired}. While we did not test this directly, we believe that the explanations written by DrugAgent are sufficiently short (an average of \textasciitilde100 words). Additionally, we believe that the use of structured JSON as part of individual agent output can also aid in interpretability.

While human expert validation would be preferable for the evaluation of all DrugAgent outputs, our evaluation framework utilizes LLM-as-a-Judge (LLMaJ), which is a scalable and cost-efficient method for evaluation of DrugAgent outputs. While deployment of LLMaJ is becoming more routine, there remain concerns about reliability and biases that may result from LLMaJ~\citep{gu2026survey,krumdick2025no}. It should be noted that these concerns are contrasted with the use of LLMaJ in the improvement of LLM performance~\citep{luo2024arena,yuan2024self}.

Similarly, our evaluation does not include systematic human expert validation of every reasoning chain or exhaustive auditing of retrieved snippets and KG paths. Because pretrained ML models and LLM-based components may encode information from prior training corpora, their training data cannot be controlled as strictly. Future work in the areas of membership inference attack research and the detection of data contamination that seeks to predict whether a particular datapoint is a member of a target model's training data may aid in developments to address this concern~\citep{duan2024membership, shokri2017membership, fu2025does}. Although it has been suggested that because pretraining typically involves relatively few passes over vast and highly diverse corpora, LLMs are generally unlikely to memorize individual training instances; our comparisons to large tabular data of assay values from Anastassiadis et al. and Tox21 sought to minimize this concern~\citep{zhang2025reviewing, anastassiadis2011comprehensive, thomas2018us}. Future development of DrugAgent with updated LLM versions needs to remain vigilant to these concerns in part by re-running stability and faithfulness checks.

\section*{Methods}

\subsection*{Overview of \mname}

\mname\ is a multi-agent architecture for biologically grounded assessment of drug-target interactions. The system integrates complementary evidence sources such as machine learning predictions, knowledge graph reasoning, and literature-derived signals, within a structured integration pipeline.

A key design principle is the separation of stochastic reasoning from deterministic decision-making. Evidence from heterogeneous sources is first organized into a structured representation and then reconciled through a rule-guided decision module, enabling consistent and interpretable outputs.

As illustrated in Fig.~\ref{fig:overview}, the system follows a modular pipeline. A Coordinator Agent manages execution, while specialized agents independently generate complementary evidence: an ML Agent produces quantitative affinity predictions, a KG agent extracts mechanistic paths, and a RAG agent retrieves literature evidence. These signals are integrated by a reasoning module that produces the final prediction and explanation.

\paragraph{Pipeline.}
The system operates in three stages: (1) coordination, which invokes specialized agents and organizes their outputs into a shared structured format; (2) evidence generation, where each agent produces a label and supporting rationale; and (3) integration, where a reasoning module synthesizes the resulting evidence and applies rule-guided fusion to produce the final decision.

\paragraph{Implementation.}
The framework is implemented using Microsoft AutoGen~\cite{wu2023autogen}. All agents use an OpenAI GPT-5.2 model via the Azure OpenAI API. Deterministic decoding (temperature = 0.0, seed = 42) is used for all components except the reasoning module. The RAG pipeline uses the \texttt{text-embedding-3-large} model for retrieval. Additional implementation details are provided in the following sections. The system prompts, structured output schemas, and rule specifications used by each agent are provided in Supplementary Section 1.1.

\subsubsection*{Coordinator Agent}

The Coordinator Agent orchestrates the overall workflow by invoking specialized agents and organizing their outputs for downstream reasoning. Each specialized agent is responsible for a distinct evidence source or reasoning component, and the Coordinator Agent ensures that their outputs are returned in a consistent, structured format. Specifically, all module outputs are constrained to a shared JSON schema, enabling heterogeneous evidence to be integrated and compared in the downstream reasoning stage.

\subsubsection*{ML Agent}

The ML Agent predicts drug-target interactions using DeepPurpose~\cite{huang2020deeppurpose}, a pre-trained model on BindingDB~\cite{liu2025bindingdb} with a message passing neural network for drug and a convolutional neural network for protein sequences. It estimates continuous $pK_d$ values from molecular SMILES strings and protein sequences. For the kinase benchmark, predicted affinities are discretized into three ordinal categories following standard conventions: Strong ($pK_d \geq 7.0$), Moderate ($6.0 \leq pK_d < 7.0$), and Weak ($pK_d < 6.0$)~\cite{ozturk2018deepdta, he2017simboost}. For the Tox21 dataset, predictions are mapped to binary labels, where Strong is treated as Active, and Moderate or Weak as Inactive, consistent with assay definitions.

\subsubsection*{KG Agent}

We construct a heterogeneous drug-target knowledge graph by integrating curated biomedical resources, including DrugBank~\citep{knox2024drugbank}, CTD~\citep{davis2023comparative}, and BindingDB~\cite{liu2025bindingdb}. Nodes represent drugs and genes, while edges encode curated drug-target interactions together with supporting evidence, including literature-derived interaction annotations (e.g., topotecan decreases TOP1 protein activity in CTD) and assay-derived measurements (e.g., topotecan-TOP1 IC50 = 21 nM in BindingDB). For each drug-target pair, textual evidence from these sources is aggregated into structured, source-aware representations, preserving interpretability across curated annotations and experimental measurements.

We perform multi-hop reasoning over this graph using breadth-first search with a maximum search depth of $D_{\max}=5$. Candidate paths are ranked using a structural scoring function that favors shorter paths while penalizing paths that pass through high-degree intermediate hub nodes. The hub penalty reduces the influence of nonspecific connectivity through broadly connected biomedical entities; a topic that has been previously considered~\citep{himmelstein2015heterogeneous,himmelstein2017systematic}.
\[
S(P) = \frac{1}{\ell(P)} \cdot \lambda^{h_{\mathrm{int}}(P)},
\]
where $\ell(P)$ is the number of edges in path $P$, and $h_{\mathrm{int}}(P)$ is the number of intermediate nodes on the path with degree greater than 300. We use $\lambda=0.6$ and retain the top-$k$ paths ($k=3$) for downstream reasoning. Sensitivity analyses showed that this configuration produced stable ranked path sets across a reasonable range of values (Supplementary Table~S5).

To characterize KG evidence availability, we computed the shortest drug-target path length for each evaluation pair using the same maximum search depth. All kinase pairs and 99.0\% of Tox21 AR pairs were connected within five hops. The kinase benchmark contained a larger fraction of direct 1-hop edges than Tox21 AR (47.0\% vs. 19.2\%), whereas Tox21 AR was dominated by indirect 3-hop paths (72.4\%; Supplementary Table~S1).

The selected paths are processed by an LLM-based component to produce a structured output consisting of a categorical label, confidence score, and supporting rationale. This design enables interpretable, mechanism-aware reasoning while reducing the influence of non-specific graph connectivity.

\paragraph{Rationale examples.}
\begin{itemize}
    \item Direct KG interaction for drug topotecan \& TOP1: ``Direct, single-edge DrugBank-supported drug–target relationship (topotecan $\leftrightarrow$ TOP1) with a specific, well-defined molecular mechanism (TOP1 poison/trapping).''
    \item Indirect KG interaction for drug topotecan \& TOP2A via copper: ``One intermediate (copper) is a strong hub and its TOP2A effects are context/compound/organism-dependent, reducing specificity; overall evidence supports an indirect, non-specific association rather than a direct mechanism.''
\end{itemize}

Compared to embedding-based link prediction methods, this path-based approach preserves explicit mechanistic interpretability, which is critical for downstream evidence integration.

\subsubsection*{PubMed RAG Agent}

We construct an unstructured evidence layer using a retrieval-augmented generation (RAG) pipeline over PubMed Central (PMC) Open Access articles retrieved in April 2026. To address sparse direct co-mentions, we employ a multi-channel retrieval strategy that issues queries at three levels: drug-target pairs, drug-only, and target-only. Pair-level retrieval used both an entity query and an assay-enriched query, whereas drug-only and target-only retrieval used assay-enriched single-entity queries. Assay-related terms included binding and target-engagement terminology such as \texttt{Kd}, \texttt{Ki}, \texttt{IC50}, \texttt{EC50}, potency, selectivity, biochemical assay, kinase assay, enzymatic activity, inhibition, phosphorylation, and target engagement. For each query type, we retrieve up to 10 documents per entity type to ensure balanced coverage across sources. The multi-channel retrieval strategy achieved broad coverage across both the kinase and Tox21 AR domains, despite substantial differences in their data distributions (Supplementary Table~S2). In the kinase dataset, retrieval yielded 722 drug-target pair-level papers, 333 drug-level papers, and 1,710 gene-level papers, whereas in the Tox21 AR dataset, retrieval yielded 795 pair-level papers, 4,223 drug-level papers, and 10 gene-level papers.

\paragraph{Retrieval and Scoring}
Query embeddings were generated using the \texttt{text-embedding-3-large} model and searched against a Facebook AI Similarity Search (FAISS) index of PMC Open Access text chunks. Retrieval was performed through three channels: drug-target pair, drug-only, and target-only. Candidate PMC articles were selected using the maximum chunk similarity score (cosine similarity) within each article, with entity filters requiring the relevant drug and/or target mention when available. If no article satisfied the entity filter, the system fell back to the highest-scoring candidates without the filter.

Within selected articles, candidate chunks were re-ranked using a fixed, label-independent score:
\[
s_{\mathrm{chunk}} =
s_{\mathrm{embed}} +
b_{\mathrm{section}} +
b_{\mathrm{entity}} .
\]
The section term favored Results passages and mildly penalized Methods passages, while the entity term prioritized drug-target co-mentions over single-entity mentions. Chunks below a minimum word-count threshold were removed. The re-ranking score above was used only to select and order chunks for the RAG prompt, not as a ground-truth activity label. Selected chunks were passed to the RAG agent with explicit retrieval-channel prefixes: \texttt{[PAIR EVIDENCE]}, \texttt{[DRUG EVIDENCE]}, or \texttt{[TARGET EVIDENCE]}. This allows the model to distinguish direct pair-level evidence from one-sided contextual evidence. Unless otherwise specified, all analyses used a fixed retrieval-scoring configuration with a Results-section bonus of $+0.5$, a Methods-section penalty of $-0.2$, a drug-target co-mention bonus of $+1.0$, a single-entity mention bonus of $+0.1$, a minimum chunk length of 15 words, and top-$k=80$ candidate chunks per query; the sensitivity ranges considered for these parameters are summarized in Supplementary Table~S6. 

In a repeated-run analysis of 100 kinase pairs, retrieved evidence quotes were fully stable across runs (mean Jaccard = 1.000 across retrieval channels), while source-level RAG labels were identical in 70\% of pairs and showed at least two-of-three agreement in 98\% (Supplementary Section~1.3).

For the Tox21 AR scoring function, the entity name (e.g., ``AR'') was expanded to include \texttt{androgen receptor}, \texttt{AR-LBD}, \texttt{AR-LBP}, and \texttt{NR3C4}, reducing ambiguity from the short gene symbol \texttt{AR}.

Separately, it should be noted that in DrugAgent, the RAG module was invoked for all drug-target pairs. The retrieved-paper count used for stratified analyses denotes direct drug-target pair-level papers only. Because RAG also performs drug-only and target-only retrieval, pairs with zero direct pair-level papers may still receive indirect literature context. Thus, the zero-paper DrugAgent stratum is not equivalent to the ML+KG ablation, where RAG is removed entirely from reasoning and final decision-making.

\subsubsection*{Reasoning Agent}

The Reasoning Agent operates in two stages: structured reasoning and rule-guided decision-making. In the first stage, outputs from the ML, KG, and RAG modules are converted into a structured JSON format that supports consistent communication and downstream reasoning. Each source contributes an \textit{evidence\_analysis} entry that includes the evidence source, agent rationale, proposed action, observations, and proposed DTI label. A summarization step then adds a \textit{summary\_reasoning} field that captures agreement, conflict, and uncertainty across sources. This stage summarizes evidence but does not produce the final decision. In the second stage, a decision module produces the final output, including a label and supporting rationale. Decisions follow a rule priority order: majority agreement is applied first, followed by tie-breaking. When these rules are inconclusive, a constrained LLM-based fallback is applied.

\subsection*{Experimental Protocol}

\subsubsection*{Evaluation design}

The primary objective was to evaluate whether DrugAgent produces evidence-aligned, internally consistent, and interpretable conclusions when integrating heterogeneous ML, KG, and literature-derived signals.
The kinase benchmark served as the primary evaluation setting. The Tox21 AR dataset was used to assess whether the observed integration behavior persisted under a distinct binary, cell-based screening regime.
To evaluate the contribution of each component, we performed ablation studies in which each module (ML, KG, or RAG) was removed from the pipeline. In each setting, the system was re-executed using only the remaining sources, and the removed component was excluded from both the reasoning stage and the final decision process. 

\subsubsection*{Diagnostics for Evidence-Grounded integration under LLM-as-a-Judge}

We evaluated DrugAgent using an LLM-as-a-Judge (LLMaJ) protocol grounded in recent analyses of evaluation reliability, bias, and faithfulness~\citep{gu2026survey,zheng2023judging,shi2025judging}. In this study, LLMaJ was used as an internal diagnostic tool rather than as an independent biological gold standard. The judge was not asked to determine whether a drug truly interacts with a target. Instead, it assessed whether each generated explanation was supported by the evidence provided to the system, whether the explanation introduced unsupported claims or contradictions, and whether the inferred association was biologically plausible given the available evidence.

This distinction is important because evidence-grounded plausibility can differ from experimental correctness, especially when retrieved literature is sparse, indirect, or conflicting. We therefore interpret LLMaJ outputs as diagnostic probes of evidence alignment, output consistency, and reasoning behavior under heterogeneous biomedical evidence, rather than as definitive validation of DTI labels or integration optimality. We operationalized these diagnostics along four complementary dimensions.

\paragraph{(1) Faithfulness.}
Faithfulness measured input-output consistency: whether the generated fusion explanation remained attributable to the provided ML, KG, and retrieved literature evidence~\citep{agarwal2024faithfulness,lyu2023faithful,gu2026survey}. 
The judge checked whether the explanation introduced unsupported claims, contradicted the provided evidence, or omitted critical evidence needed to justify the stated conclusion.
Thus, faithfulness serves as an evidence-grounded hallucination check rather than a measure of biological correctness or integration quality.
Faithfulness was scored on a five-point scale, where 1 indicated highly unfaithful and 5 indicated fully faithful.
The judge also produced binary indicators for grounding, contradiction, unsupported claims, and critical omissions, together with a brief rationale.
Because this metric is conditioned on the evidence available to each system variant, high faithfulness is expected for constrained structured-output systems and should not be interpreted as evidence of better multiple source integration.

\paragraph{(2) Biological Plausibility.}
Biological plausibility evaluated whether the inferred drug-target association could be interpreted as a reasonable biological or pharmacological hypothesis given the provided evidence~\citep{petridis2026holistic}. As input, the judge receives a combination of the drug name, gene name, and DrugAgent's generated final output (i.e., evidence summary). The judge generates three outputs: 1) a five-point plausibility score, 2) a binary indicator of whether the association was contradicted by the input, and 3) a binary indicator of whether the rationale was mechanistically coherent. The plausibility score ranged from 1 to 5, with higher scores indicating that the judge rated the association as more biologically plausible and mechanistically coherent given the provided evidence.

Examples of contradiction include 1) A binds B/A does not bind B or 2) A strongly binds B/A weakly binds B. Similarly, mechanistic coherence indicates whether the association could reasonably be considered a possible biological hypothesis based on the input; sources of incoherence can arise from poor supporting evidence and lack of support for direct interaction. The judge also provided a brief rationale for each assessment. The full judge prompts, scoring definitions, and output schemas are provided in Supplementary Section 1.2. As with faithfulness, plausibility was interpreted as a diagnostic proxy rather than as experimental validation of biological correctness.

\paragraph{(3) Stability.}
Stability evaluated the reproducibility of integration outcomes under repeated executions.
For the kinase benchmark, we randomly sampled $N=100$ cases and ran the full integration pipeline three times ($k=3$).
We report label stability as the fraction of cases for which all runs yielded identical categorical outputs:
\[
\text{AgreementRate} = \frac{1}{N}\sum_{i=1}^{N}\mathbf{1}\{\hat{y}_i^{(1)}=\hat{y}_i^{(2)}=\hat{y}_i^{(3)}\}.
\]
When generated rationales were compared across repeated runs, we additionally measured reasoning stability using embedding-based semantic similarity (i.e., cosine similarity) between explanation outputs.

\paragraph{(4) Evidence Combination Consistency and Conflict Dynamics.}
For the kinase benchmark, we analyzed integration behavior conditioned on combinations of module outputs (ML, KG, RAG), representing each case as a triplet of predicted activity labels, such as W|M|S (W: weak, M: moderate, and S: strong).
For each combination, we compared the distribution of ground-truth labels and aggregated outputs and examined label transitions from module predictions to the final decision.
This analysis was intended to distinguish systematic integration behavior from raw predictive performance and to identify whether conflict resolution was driven by consensus, majority evidence, or a shift toward intermediate activity labels under conflicting evidence; this fallback-to-intermediacy behavior has been observed before~\cite{pitre2025consensagent}.

\subsection*{Classification Evaluation}

To quantify predictive consistency, we computed accuracy and F1 score. We compared individual modules (ML, KG, RAG), ablation variants (KG+RAG, ML+RAG, ML+KG), and the full DrugAgent model.

\paragraph{Leakage Control.}
Ground-truth interaction labels were not exposed to any agent during integration. Retrieval was restricted to PMC Open Access articles and curated knowledge graph sources. The LLMaJ was provided only with raw agent outputs as JSON and final explanations, ensuring independence from ground-truth labels. For the Tox21 Androgen Receptor (AR) dataset, literature availability (i.e., publication counts) was used only to ensure sufficient retrievable evidence for evaluation and was not used to assign activity labels. We did not perform a temporal or entity-level split that explicitly removes evaluation drug-target pairs from the curated KG or retrieved literature corpus. Thus, KG and RAG modules should be interpreted as evidence-access components rather than as predictors evaluated under a strict unseen-interaction generalization setting.

\clearpage
\bibliography{sample}

\section*{Data availability statement}

The kinase inhibition dataset and Tox21 assay data are publicly available from the original studies. Knowledge graph evidence was derived from BindingDB, DrugBank, and CTD obtained in April 2026, and literature evidence was retrieved from PubMed Central Open Access articles. Processed datasets and code are available at the project repository or from the corresponding author upon reasonable request, where redistribution restrictions apply. Sample outputs from DrugAgent are on GitHub repository~\url{https://github.com/sciluna/DrugAgent}.

\section*{Funding}

This research was partly supported by the Division of Intramural Research (DIR) of the National Library of Medicine (NLM), National Institutes of Health (NIH) (ZIALM240126). Tianfan Fu is supported by the Young Scientists Fund (C Class) of the National Natural Science Foundation of China (Grant No. 62506154), the Fundamental Research Funds for the Central Universities, and Nanjing University International Collaboration Initiative (Grant No. 020214380129), and the ``111 Center''(3).

\section*{Competing Interests}

The authors declare no competing interests.

\section*{Author contributions statement}

T.F., Y.I., and A.L. conceived the original idea. Y.I., T.S., and A.L. conceived the experiment(s). Y.I., T.S., and X.W. conducted the experiment(s), and Y.I., T.S., X.W., R.K., and A.L. analysed the results. Y.I., T.F., T.S., R.K., and A.L. wrote the manuscript draft. All authors reviewed the manuscript.

\end{document}